\documentclass[twoside]{article}

\usepackage[accepted]{aistats2020}
\usepackage[round]{natbib} 

\usepackage{algorithm,algpseudocode}
\usepackage{graphicx}
\usepackage{YK} 

\usepackage{enumitem} 
\usepackage{hyperref} 
\hypersetup{
    colorlinks = true,
    allcolors = Blue
}

\begin{document}

%

%

\twocolumn[

\aistatstitle{Auditing ML Models for Individual Bias and Unfairness}

\aistatsauthor{ Songkai Xue \And Mikhail Yurochkin \And Yuekai Sun }

\aistatsaddress{Department of Statistics \\ University of Michigan \And IBM Research \\ MIT-IBM Watson AI lab \And Department of Statistics \\ University of Michigan} ]


\begin{abstract}
We consider the task of auditing ML models for individual bias/unfairness. We formalize the task in an optimization problem and develop a suite of inferential tools for the optimal value. Our tools permit us to obtain asymptotic confidence intervals and hypothesis tests that cover the target/control the Type I error rate exactly. To demonstrate the utility of our tools, we use them to reveal the gender and racial biases in Northpointe's COMPAS recidivism prediction instrument.
\end{abstract}

\section{Introduction}

Machine learning (ML) models are finding their way into high-stakes decision making tasks such as housing \citep{angwin2016Facebook,angwin2017Facebook} and recidivism prediction \citep{angwin2016Machine}. Although replacing humans with ML models eliminates human biases in the decision-making process, the models may perpetuate or even exacerbate biases in their training data. Such biases in ML systems are especially objectionable if they adversely affect minority and/or underprivileged groups of users \citep{barocas2016Big}. For example, in 2016 and 2017, ProPublica reported that Facebook allows advertisers to filter users by attributes protected by federal anti-discrimination law \citep{angwin2016Facebook,angwin2017Facebook}. Similar reports eventually prompted state and federal level investigations into Facebook's advertising platform \citep{tobin2019HUD,tobin2019New}. Other high-profile examples of algorithmic bias/unfairness include racial bias in algorithms for estimating defendants' chances of committing another crime \citep{angwin2016Machine}, gender biases in resume screening systems for technical positions \citep{dastin2018Amazon}, and racial bias in image search results \citep{allen2016three}.

In response, the data science community has proposed many formal definitions of algorithmic fairness and methods to train ML models that abide by the definitions. However, a notable gap in the literature remains: \emph{calibrated} methods for detecting and localizing bias/unfairness in ML models. For example, in the aforementioned investigations of bias/unfairness in ML models, investigator study discrepancies between summary statistics of the output of ML models on subgroups (\eg\ false positive rates on black and white defendants) \citep{angwin2016Machine, dastin2018Amazon}, but they lack statistical tools to ascertain whether the discrepancies they observe are systemic or due to the inherent randomness in the data. In other words, the investigators lack tools to calibrate the statistics so that the chance of a false alarm is controlled. 

In this paper, we address this issue by providing a suite of inferential tools for detecting and localizing bias/unfairness in ML models. The main benefits of the methods are
\begin{enumerate}
\item the methods only require \emph{black-box} or query access to the ML model: an auditor only has to observe the output of the ML model;
\item the methods are \emph{computationally efficient}: the main computational expense is solving a convex optimization problem;
\item the methods provide an \emph{interpretable} pairing between inputs that localize the bias/unfairness in an ML system.
\end{enumerate} 
The basis of the proposed suite of inferential tools is a result on the asymptotic distribution of the optimal value of a convex optimization problem. Due to the lack of regularity in the value function of the problem, the asymptotic distribution of the optimal value is non-Gaussian. This result may be of independent interest to researchers.

\subsection{Related work}

Generally speaking, there are two kinds of mathematical definitions of algorithmic fairness: group fairness and individual fairness. Most prior work on algorithmic fairness focuses on group fairness because it is suitable for statistical analysis. Despite its prevalence, group fairness suffers from two critical issues. First, it is possible for an ML model that satisfies group fairness to be blatantly unfair from the perspective of individual users \citep{dwork2012fairness}. Second, there are fundamental incompatibilities between intuitive notions of group fairness \citep{kleinberg2016Inherent,chouldechova2017Fair}. 

In light of the issue with group fairness, we focus on individual fairness in this paper. At a high-level, the idea of individual fairness is a fair algorithm ought to treat similar users similarly. This idea is intuitive and has a strong legal basis. Despite its benefits, individual fairness has been dismissed as impractical because there is no consensus on which users are similar. Although this is a critical issue, it is not the focus of this paper, and we assume there is a \emph{similarity function} that determines which users are similar and which users are dissimilar in the rest of the paper. Our tools make no restrictions on the similarity function, so auditors are free to customize the similarity function for their applications. In our computational results, we follow \cite{yurochkin2020Training} by adopting a data-driven similarity function.

There is a parallel vein of work in Wasserstein distributionally robust optimization (DRO) \citep{blanchet2019quantifying,lee2018minimax,sinha2017Certifying,blanchet2019Confidence} on obtaining confidence intervals for the population optimal value. The latest in this line of work \citep{blanchet2019Confidence} also obtains asymptotic distributional results on the distributionally robust optimal value. The key distinction between this line of work and our work is the robustness radius $\varepsilon$ is fixed in our work and shrinking (usually at a $\frac1n$-rate) in the DRO literature. As we shall see, this leads to qualitatively different distributional results: the asymptotic distribution under a fixed radius is generally non-Gaussian, while the distribution under a shrinking radius is Gaussian.

\section{The auditor's problem}
\label{sec:auditorsProblem}

Imagine an investigator evaluating the fairness of an ML model. The auditor wishes to detect and localize violations of \emph{individual fairness} in the ML models. In this section, we formalize the auditor's task in a convex optimization problem. We start by recalling the definition of individual fairness by \cite{dwork2012fairness}.

\begin{definition}
\label{def:individualFairness}
An ML model $h:\cX\to\cY$ is individually fair if there is $L > 0$ such that
\[
d_y(h(x_1),h(x_2)) \le L d_x(x_1,x_2)\text{ for all }x_1,x_2\in\cX,
\]
where $d_x$ and $d_y$ are metrics on the input space $\cX$ and the output space $\cY$.
\end{definition}

The fair metric $d_x$ in Definition \ref{def:individualFairness} encodes our intuition of which samples should be treated similarly by the ML model. We emphasize that $d_x(x_1,x_2)$ being small does NOT imply $x_1$ and $x_2$ are similar in all respects. Even if $d_x(x_1,x_2)$ is small, $x_1$ and $x_2$ may differ in certain attributes that are irrelevant to the ML task at hand, \eg, protected attributes. 

At a high-level, we envision the auditor collects a set of audit data and evaluates the performance of the ML model on the audit data and checks for discrepancies between the performance of the model on similar samples. The presence of large discrepancies suggests the ML model violates individual fairness. This type of audit is known as a \emph{correspondence study} in the empirical literature in social sciences; \cite{bertrand2004Are}'s celebrated study of discrimination in the US labor market is a prominent example.

\paragraph{Mathematical preliminaries} Denote the input and output space of the ML model by $\mathcal{X}$ and $\mathcal{Y}$ respectively and the sample space by $\mathcal{Z} \triangleq \mathcal{X} \times \mathcal{Y}$. We equip $\mathcal{X}$ with a metric $d_x: \mathcal{X} \times \mathcal{X} \to \mathbb{R_+}$. This metric is the metric appearing in Definition \ref{def:individualFairness}; it encodes our intuition of which samples are similar and which are dissimilar. To keep things simple, we assume $\cY$ is a discrete set (\ie\ the ML model is a classifier). We equip $\mathcal{Z}$ with the metric 
\begin{equation*}
d_z((x_1,y_1),(x_2,y_2)) \triangleq d_x(x_1, x_2) + \infty \times \boldsymbol{1}\{y_1 \neq y_2 \}, 
\end{equation*}
The metric $d_z$ encodes our intuition of which samples are similar and which are dissimilar: $(x_1,y_1)$ and $(x_2,y_2)$ similar if and only if (i) they share a label and (ii) $x_1$ and $x_2$ are similar according to $d_x$. Finally, we equip $\Delta(\mathcal{Z})$, the set of probability distributions on $\mathcal{Z}$, with the 1-Wasserstein distance. Recall the Wasserstein distance between two probability distributions $P$ and $Q$ on $\mathcal{Z}$ is
\[
W(P, Q)=\inf _{\Pi \in \mathcal{C}(P, Q)} \int_{\mathcal{Z} \times \mathcal{Z}} c\left(z_{1}, z_{2}\right) d \Pi\left(z_{1}, z_{2}\right),
\]
where $c:\mathcal{Z}\times\mathcal{Z}\to\reals_+$ is a transportation cost function and $\mathcal{C}(P, Q)$ is the set of couplings between $P$ and $Q$. To encode our intuition of fairness in the Wasserstein distance, we use $d_z^2$ as the transportation cost function. This Wasserstein distance considers two distributions close if the mass they put on comparable segments of the sample space is similar (the placement of mass within comparable segments may differ). 

Returning to the auditor's task, let $h$ be the ML model under audit. To detect and localize disparate treatment by the ML model, the auditor picks a loss function $\ell_h:\cZ\to\reals_+$ to measure the performance of the model and evaluates the risk of the model $\mathbb{E}_{Z\sim P_\star}[\ell_h(Z)]$, where $P_\star$ is the data generating distribution. If there is no bias/unfairness in the ML model, then it is not possible for the auditor to increase the risk by moving (probability) mass to similar areas of the sample space. In other words, if the ML model is fair, then the value of the optimization problem \vspace{-0.25em}
\begin{equation}
\label{eq:populationAuditorsProblem}
\begin{aligned}
& \max_{P \in \Delta(\mathcal{Z})} & &\mathbb{E}_{Z\sim P}[\ell_h(Z)] - \mathbb{E}_{Z\sim P_\star}[\ell_h(Z)] \\
& \operatorname{subject~to} & & W(P, P_\star) \leq \varepsilon,
\end{aligned}\vspace{-0.25em}
\end{equation}
where $\varepsilon \geq 0$ is a transportation budget parameter and should be small. The constraint on the transportation budget compels the auditor to move mass to similar areas of the sample space. 

In practice, $P_\star$ is unknown, so the auditor collects a set of audit data $\{(x_i, y_i)\}_{i = 1}^n$ and solves the empirical version of \eqref{eq:populationAuditorsProblem}: \vspace{-0.25em}
\begin{equation}
\begin{aligned}
& \max_{P \in \Delta(\mathcal{Z})} & &\mathbb{E}_{Z\sim P}[\ell_h(Z)] - \mathbb{E}_{Z\sim P_n}[\ell_h(Z)] \\
& \operatorname{subject~to} & & W(P, P_n) \leq \varepsilon,
\end{aligned}\vspace{-0.25em}
\label{eq:auditorsProblem}
\end{equation}
where $P_n$ is the empirical distribution of the audit data. A large optimal value is evidence that the ML model is unfair. This suggests the optimal value of this optimization problem as a test statistic. We call the optimal value of \eqref{eq:auditorsProblem} the Fair Transport Hypothesis (FaiTH) test statistic. In summary, if the ML model is fair, then the FaiTH statistic is small.


The FaiTH statistic is robust to small changes in the similarity functions. Let $d_x, d_{x_*}: \cX \times \cX \to \bbR_+$ be two different similarity metrics on $\cX$. Let $c, c_*: \cZ \times \cZ \to \bbR_+$ be the transportation cost functions on $\cZ$ induced by $d_x, d_{x_*}$. Let $W, W_*: \Delta(\cZ) \times \Delta(\cZ) \to \bbR_+$ be the Wasserstein distances on $\Delta(\cZ)$ induced by $d_x, d_{x_*}$. We start by stating the following assumptions:

\begin{enumerate}[label={(A\arabic*)}, noitemsep,topsep=0pt,leftmargin=*]
    \item the feature space $\cX$ is bounded: \begin{equation*}
        D \triangleq \max\{\diam(\cX), \diam_*(\cX)\} < \infty;
    \end{equation*}
    \item the loss function is non-negative and bounded: $0\leq \ell_h(z)\leq M$ for all $z \in \cZ$, and $L$-Lipschitz with respect to $d_x$ and $d_{x_*}$:
    \begin{equation*}
    \begin{aligned}
        &\textstyle\sup_{y:(x_1, y), (x_2, y) \in \cZ} \left|\ell_h(x_1, y) - \ell_h(x_2, y)\right| \\
        &\quad\leq Ld_x(x_1, x_2)\wedge d_{x_*}(x_1, x_2);
    \end{aligned}
    \end{equation*}
    \item the discrepancy between the transportation cost functions is uniformly bounded:
    \begin{equation*}
        \sup_{(x_1, y), (x_2, y) \in \cZ} \left|\begin{aligned}c((x_1,y), (x_2,y)) - \\ c_*((x_1,y), (x_2,y)) \end{aligned}\right| \leq \eta D^2.
    \end{equation*}
\end{enumerate}


The following proposition shows the robustness of the FaiTH statistic with respect to changes in the similarity functions.
\begin{proposition}\label{prop:robustFaiTH}
Under Assumptions A1--A3, the difference between the FaiTH statistics induced by $d_x$ and $d_{x_*}$ satisfies \vspace{-0.3em}
\begin{equation*}
    \left|\begin{aligned}\textstyle\max_{P: W(P, P_n) \leq \varepsilon} \bbE_{Z \sim P}[\ell_h(Z)] - \\ \textstyle\max_{P: W_*(P, P_n) \leq \varepsilon} \bbE_{Z \sim P}[\ell_h(Z)] \end{aligned}\right| \leq \frac{L\eta D^2}{\sqrt{\varepsilon}}. \vspace{-0.3em}
\end{equation*}
\end{proposition}

In the subsequent sections, we develop a suite of inferential tools based on the FaiTH statistic. We emphasize that 
\begin{enumerate}[noitemsep,topsep=0pt]
\item the auditor only needs to be able to query the output of the ML model to collect the audit data;
\item \eqref{eq:auditorsProblem} is a linear program, so it is possible to evaluate the FaiTH statistic efficiently.
\end{enumerate}


Inference for the optimal value of an optimization problem \eqref{eq:auditorsProblem} is generally a hard task, and we focus on finite sample spaces. This simplification is common in the literature on inferential tools for optimal transport problems \citep{sommerfeld2018inference,klatt2018Empirical}. As we shall see, the restriction of finite
spaces is sufficient for many practical problems, including evaluating the algorithmic fairness of the COMPAS recidivism prediction instrument. For a finite sample space, the auditor's problem is \vspace{-0.3em}
\[
\begin{aligned}
&\max_{\Pi\in\reals_+^{|\cZ|\times|\cZ|}} & &l^\top(\Pi^\top \boldsymbol{1}_{|\cZ|} - f_n) \\
&\operatorname{subject~to}                        & & \langle C,\Pi\rangle \le \varepsilon \\
&                                  & & \Pi \boldsymbol{1}_{|\cZ|} = f_n,
\end{aligned} \vspace{-0.3em}
\]
where $l\in\reals_+^{|\cZ|}$ is the vector of losses and its $i$-th entry is $\ell_h(z_i)$, $C\in\reals_+^{|\cZ|\times|\cZ|}$ is the matrix of transportation costs and its $(i,j)$-th entry is $c(z_i,z_j)$, and $f_n \in\Delta_{|\cZ|}$ is the empirical distribution of the data $\{(x_i, y_i)\}_{i = 1}^n$. 

\section{Asymptotic distribution of the FaiTH statistic}
\label{sec:asymptoticDistribution}

In this section, we establish our main result on the asymptotic distribution of the FaiTH statistic. We state the main result and provide a sketch of the proof. For completeness, we also describe the key ingredients of the proof along the way.

\subsection{Asymptotic distribution}

The sample space of our interest is discrete: $\mathcal{Z} = \{z_1, \cdots, z_K\}$, where $K = |\mathcal{Z}|$, and the data generating distribution is $P_\star = \sum_{k=1}^K f_{\star}^{(k)} \delta_{z_i}$, where $f_\star = (f_\star^{(1)}, \cdots, f_\star^{(K)})^\top \in \Delta_K \triangleq \{x\in\reals_+^K:\boldsymbol{1}_K^\top x = 1\}$ and $\delta_z$ is the Dirac measure at $z$. The auditor observes an empirical measure $P_n = \sum_{k=1}^K f_{n}^{(k)} \delta_{z_i}$ based on frequency summary of IID samples $Z_1, \cdots, Z_n \sim P_\star$, \ie, $f_n^{(k)} = |\{i\in [n]: Z_i = z_k \}|/ n$ for $k = 1, \cdots, K$, and $f_n = (f_n^{(1)}, \cdots, f_n^{(K)})^\top \in \Delta_K$. Hereafter, we do not distinguish between measures $P_\star, P_n$ and their corresponding probability vectors $f_\star, f_n$.

Consider the audit value function $\psi:\Delta_K\to\reals_+$ defined as \vspace{-0.3em}
\begin{equation}\label{eq:auditValueFunction}
\begin{aligned}
\psi(f) \triangleq && & \max_{\Pi\in \mathbb{R}_{+}^{K\times K}} & &l^\top (\Pi^\top \boldsymbol{1}_K - f) \\
&& & \operatorname{subject~to} & & \langle C,\Pi\rangle \leq \varepsilon \\
&& & & & \langle D,\Pi\rangle = 0 \\
&& & & &\textstyle\Pi \boldsymbol{1}_K = f
\end{aligned} \vspace{-0.3em}
\end{equation}
where $C \in \mathbb{R}_+^{K\times K}$ is the cost matrix, $D \in \{0, 1\}^{K\times K}$ is the indicator matrix. The FaiTH statistic is the optimal value $\psi(f_n)$. The second constraint $\langle D,\Pi\rangle = 0$ explicitly encodes any restrictions on the transportation plan implicit in the transportation cost function. If $D_{i,j} = 1$, then moving mass from $z_i$ to $z_j$ is prohibited. This is equivalent to $c(z_i,z_j) = \infty$.

\begin{theorem}[Asymptotic distribution of the FaiTH statistic]\label{thm:asymptoticDistribution}
Let $f_\star \in \Delta_K$ and $nf_n \sim \operatorname{Multinomial}(n; f_\star)$. Let $l = (l_1, \cdots, l_K) \in \mathbb{R}_+^K, \varepsilon \geq 0$, $C \in \mathbb{R}_+^{K \times K}$, and $D \in \{0,1\}^{K \times K}$. Define the set \vspace{-0.3em}
\begin{equation}\label{eq:LambdaSet}
\begin{aligned}
    \Lambda = \underset{\nu, \mu \geq 0, \lambda \in \mathbb{R}^K}{\argmax} \{& \varepsilon \nu + f_\star^\top \lambda: \\
    &\nu C + \mu D + \lambda \boldsymbol{1}_n^\top \preceq_{\mathbb{R}_+^{K \times K}} - \boldsymbol{1}_n l^\top\} \vspace{-0.3em}
\end{aligned}
\end{equation}
and the multinomial covariance matrix \vspace{-0.3em}
\begin{equation*}
    (\Sigma(p))_{i,j} = \begin{cases}
        p_i(1-p_i), & \text{if}~1\leq i = j\leq K;\\
        -p_i p_j, & \text{if}~1\leq i \neq j\leq K.
        \end{cases} \vspace{-0.3em}
\end{equation*}
The asymptotic distribution of $\psi(f_n)$ is \vspace{-0.3em}
\[
    \sqrt{n}\{\psi(f_n) - \psi(f_\star)\} \stackrel{d}{\rightarrow} \inf \{ (\lambda + l)^\top Z: (\nu, \mu,\lambda) \in \Lambda \}, \vspace{-0.3em}
\]
where $Z \sim \mathcal{N} (\boldsymbol{0}_{K},\Sigma(f_\star))$.
\end{theorem}

\begin{figure}
    \vspace{-0.05in}
    \centering
    \includegraphics[width=0.23\textwidth]{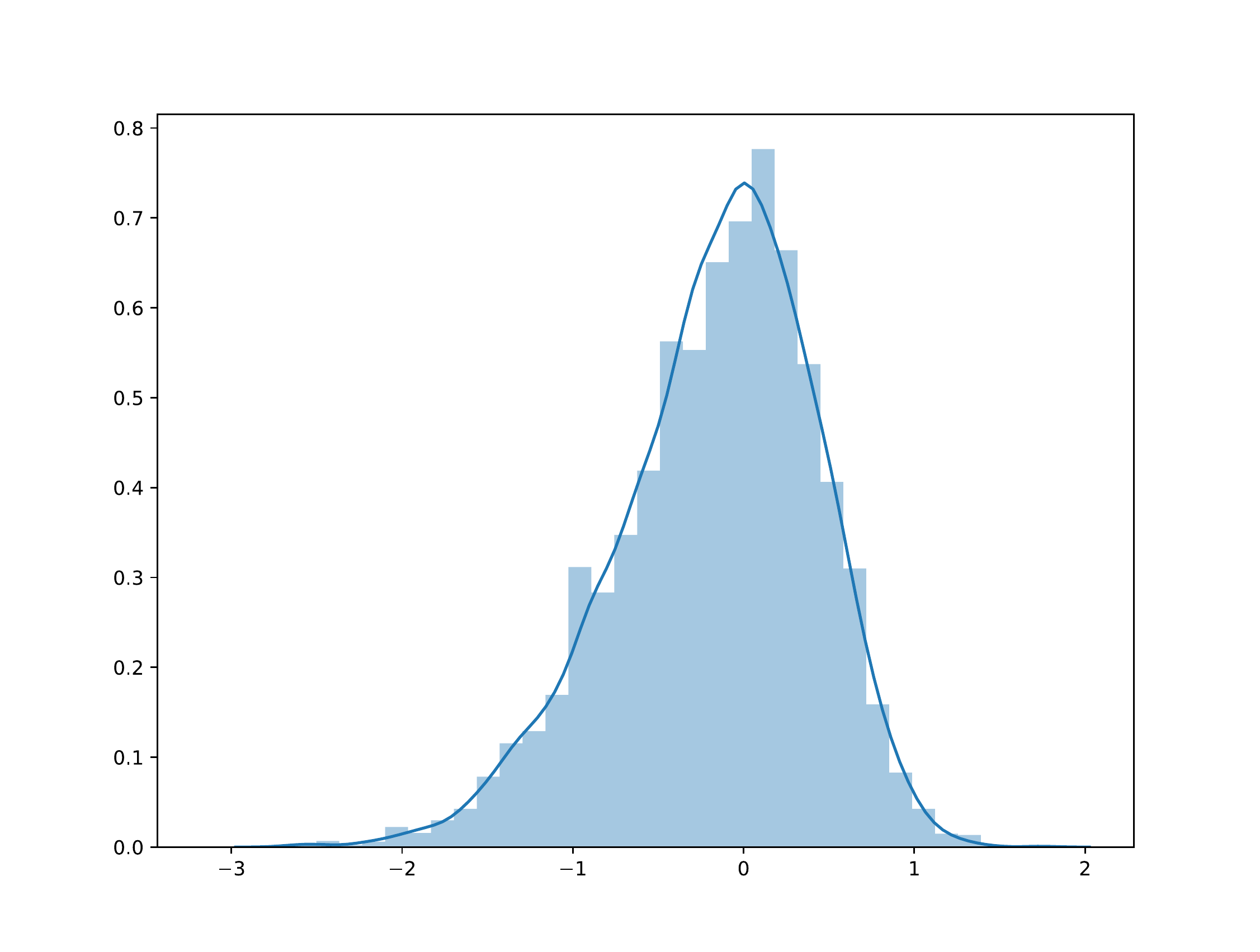} 
    \includegraphics[width=0.23\textwidth]{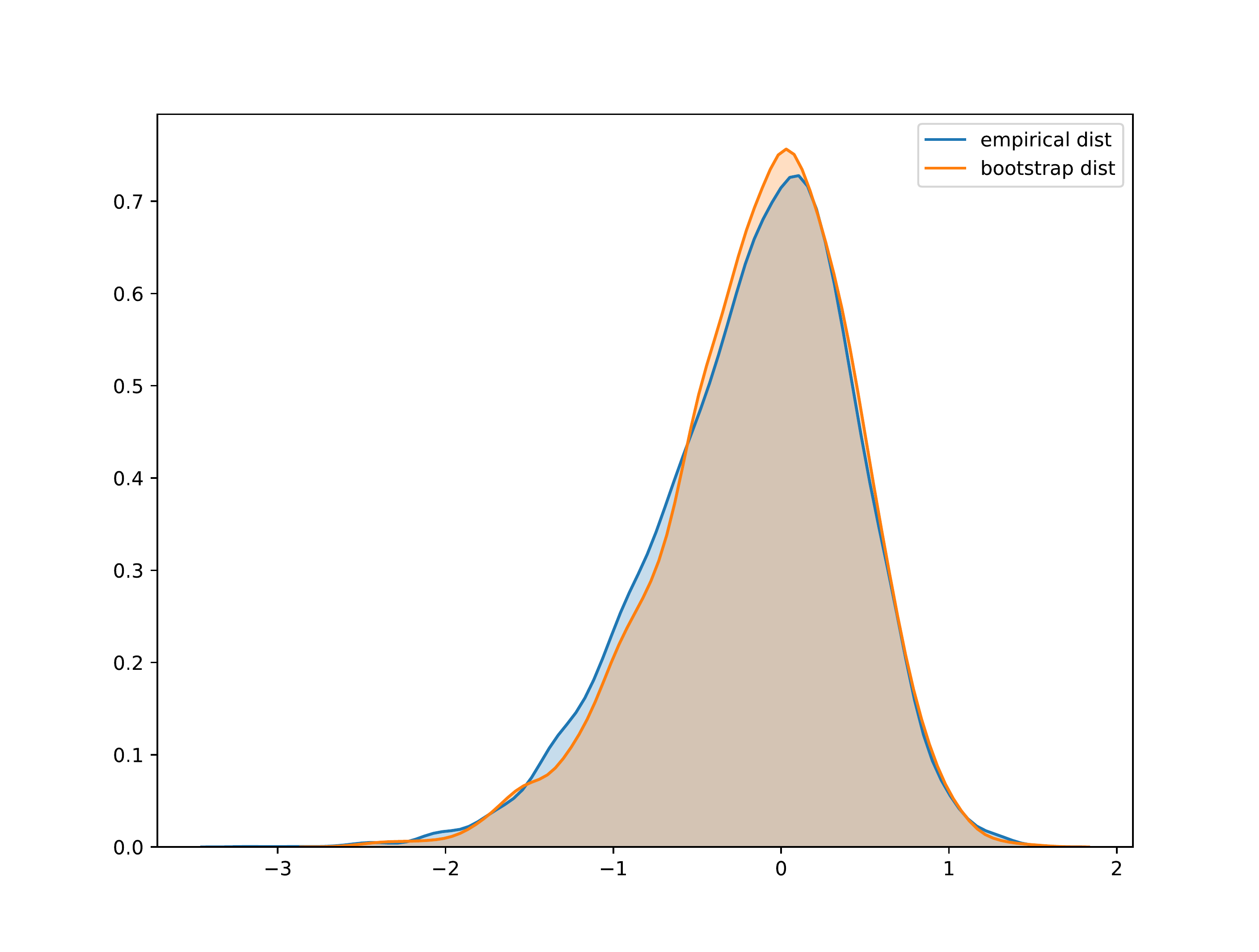}
    \caption{Asymptotic approximation (left panel) and bootstrap approximation (right panel) to the sampling distribution of the FaiTH statistic.}
    \label{fig:asymptoticDistribution}
    \vspace{-0.15in}
\end{figure}

The set $\Lambda$ in Theorem \ref{thm:asymptoticDistribution} is the set of optimal points of the dual problem of $\psi(f_\star)$, which coincides with the set of Lagrange multipliers of $\psi(f_\star)$ satisfying the optimality conditions. It is generally a convex set. However, if $\Lambda$ is a singleton, then the asymptotic distribution is Gaussian. This is the generic case, as the inequality constraint in the auditor's problem is generally active. The dual optimum is only non-unique when the inequality constraint is redundant. The left panel of Figure \ref{fig:asymptoticDistribution} shows a histogram of the values of $\sqrt{n}\{\psi(f_n) - \psi(f_\star)\}$ and its asymptotic distribution.

\subsection{Directionally differentiable statistical functionals and delta method}

A standard tool for deriving the asymptotic distribution of a statistical functional is the delta method. However, the delta method requires the statistical functional to be differentiable \citep{vandervaart1998Asymptotic}. Although the audit value function is not differentiable, it is convex and directionally differentiable. As we shall see, this allows us to appeal to a version of the delta method for directionally differentiable functions.

\begin{definition}[Hadamard directional derivatives]
$\mathbb{D}$ and $\mathbb{E}$ are Banach spaces. A map $\phi:\mathbb{D}_{\phi} \subseteq \mathbb{D} \rightarrow \mathbb{E}$ is called Hadamard directionally differentiable at $\theta_0 \in \mathbb{D}$ tangentially to $\mathbb{D}_0 \subseteq \mathbb{D}$ if there is a map $\phi^\prime_{\theta_0}: \mathbb{D} \to \mathbb{E}$ such that \vspace{-0.3em}
\[\textstyle
    \lim_{h^\prime \to h, t \to 0^+}  \frac1t(\phi(\theta_0 + t h^\prime) - \phi(\theta_0)) = \phi^\prime_{\theta_0} (h) \vspace{-0.3em}
\]
for any $h \in \mathbb{D}_0$
\end{definition}

The audit value function is closely related to the \emph{optimal value function} of the auditor's problem. The optimal value function describes the sensitivity of the optimal value of an optimization problem to perturbations of the problem parameters. Under suitable conditions, the optimal value function is directionally differentiable. 


There is a more general version of the delta method for directionally differentiable statistical functionals \citep{shapiro1991asymptotic, dumbgen1993nondifferentiableSK, romisch2014delta}. Although this version is common in the stochastic optimization literature, it rarely appears in the statistics literature. 

\begin{theorem}[Delta method]\label{thm:deltaMethod}
Suppose the following assumptions hold:
\begin{enumerate}[noitemsep,topsep=0pt,leftmargin=*]
    \item $\mathbb{D}$ and $\mathbb{E}$ are Banach spaces;
    \item $\phi : \mathbb{D}_{\phi} \subseteq \mathbb{D} \rightarrow \mathbb{E}$ is Hadamard directionally differentiable at $\theta_0$ tangentially to $\mathbb{D}_0$;
    \item $\theta_{0} \in \mathbb{D}_{\phi}$ and $\hat{\theta}_{n} :\left\{X_{i}\right\}_{i=1}^{n} \rightarrow \mathbb{D}_{\phi}$ satisfies $r_{n}\{\hat{\theta}_{n}-\theta_{0}\} \stackrel{d}{\rightarrow} \mathbb{G}_{0}$ in $\mathbb{D}$ for some $r_{n} \uparrow \infty$;
    \item $\mathbb{G}_0$ is tight and its support is included in $\mathbb{D}_{0}$.
\end{enumerate}
Then, we have \vspace{-0.5em}
\begin{equation*}
    r_{n}\{\phi(\hat{\theta}_{n})-\phi(\theta_{0})\} \stackrel{d}{\rightarrow} \phi_{\theta_{0}}^{\prime}(\mathbb{G}_{0})\text{ in }\mathbb{E}.
\end{equation*}
\end{theorem}

\subsection{Proof sketch of Theorem \ref{thm:asymptoticDistribution}}


Since $\psi(f)$ can be viewed as the optimal value function of a class of maximization problems parameterized by $f$, we can show $\psi(f)$ is Hadamard directionally differentiable at $f_\star$, and give an exact derivative formula by using Proposition 4.27 in \cite{bonnans2000Perturbation}.

\begin{theorem}\label{thm:HadamardDerivative}
Under the same assumptions of Theorem \ref{thm:asymptoticDistribution}, $\psi(f)$ is Hadamard directionally differentiable at $f_\star$. Furthemore, the derivative is given by \vspace{-0.3em}
\begin{align*}
    \psi_{f_\star}^\prime(h) &= \lim_{\substack{h^\prime \to h \\ t \to 0^+}} \frac{\psi(f_\star + th^\prime) - \psi(f_\star)}{t} \\
    &= \inf \{(\lambda + l)^\top h: (\nu, \mu, \lambda) \in \Lambda\}, \vspace{-0.3em}
\end{align*}
where the convex set $\Lambda$ is defined by \eqref{eq:LambdaSet}.
\end{theorem}

With Theorem \ref{thm:HadamardDerivative}, we can directly show the asymptotic distribution result by applying delta method for Hadamard directionally differentiable functionals.





\section{Testing whether an ML model is fair}
\label{sec:fairTesting}

Theorem \ref{thm:asymptoticDistribution}, while insightful, is not immediately useful for inference because the asymptotic distribution depends on the unknown $f_*$. In this section, we show that a bootstrap approximation to the asymptotic distribution is valid, so it is possible to perform inference with the bootstrap. Due to the non-differentiability of the audit value function \eqref{eq:auditValueFunction}, Efron's non-parametric boostrap \citep{efron1979bootstrap} is generally invalid. Instead, we consider $m$-out-of-$n$ bootstrap \citep{dumbgen1993nondifferentiableSK} and a numerical bootstrap \citep{hong2018numerical,hong2020numerical}.

\subsection{Boostrapping the asymptotic distribution of the FaiTH statistic}

We start by describing the failure of Efron's non-parametric bootstrap. Let $f_n^*$ be the empirical distribution of $n$ independent samples from $f_n$. The non-parametric bootstrap approximates the distribution of the FaiTH statistic with the distribution of $\sqrt{n}(\psi(f_n^*) - \psi(f_n))$. This distribution is known as the bootstrap distribution, and the non-parametric bootstrap is consistent if the bootstrap distribution converges weakly to the asymptotic distribution: \vspace{-0.3em}
\[
\sup _{g\in \operatorname{BL}_1 (\mathbb{R})}\left|\begin{array}{c}{\bbE^{*}\left[ g\left(\sqrt{n}\left\{\psi(f_{n}^*) - \psi(f_n) \right\}\right) | f_n\right]} \\ {-\bbE\left[g\left(\sqrt{n}\left\{\psi(f_n) - \psi(f_\star) \right\} \right) \right]}\end{array}\right| \stackrel{p}{\rightarrow} 0,\vspace{-0.3em}
\]
where $\operatorname{BL}_1 (\mathbb{R})$ is 1-Lipschitz subset of the $\|\cdot\|_\infty$ ball. Unfortunately, if $\psi$ is only directionally differentiable (but not differentiable), then the non-parametric bootstrap may fail \citep{bickel2012Resampling,andrews2000Inconsistency}. In fact, it is known that if $\sqrt{n}(f_n - f_*)$ has a Gaussian asymptotic distribution, then the non-parametric bootstrap is consistent if and only if $\psi$ is (Hadamard) differentiable \citep{fang2019inference}. Unfortunately, as saw in Section \ref{sec:asymptoticDistribution}, the FaiTH statistic is a generally non-differentiable function of the empirical distribution. 

Before discussing alternatives to the non-parametric bootstrap, we observe that the audit value function is differentiable at $f_*$ whenever $\Lambda$ is a singleton. In such problems, $\sqrt{n}(f_n - f_*)$ has a Gaussian asymptotic distribution, so the non-parametric bootstrap is consistent. One practical heuristic to check for failure of the non-parametric bootstrap is checking whether the bootstrap distribution is Gaussian: non-Gaussianity suggests failure of the non-parametric bootstrap.

Fortunately, there are several alternatives to the non-parametric bootstrap that remain consistent for non-differentiable statistical functionals. We refer to these methods as non-standard bootstrap methods. Three promiment methods are the $m$-out-of-$n$ bootstrap \citep{dumbgen1993nondifferentiableSK,shao1994Bootstrap,bickel2008choice}, subsampling \citep{politis1999subsampling}, and the numerical bootstrap \citep{hong2018numerical,hong2020numerical}. In our computational results, we rely on the $m$-out-of-$n$ bootstrap and the numerical bootstrap. We provide detailed descriptions of both methods in Section \ref{sec:bootstrapMethods} of the Supplementary Materials.



\begin{theorem}[Consistency of $m$-out-of-$n$ bootstrap]\label{thm:bootConsistency1}
Let $m f_{n,m}^* \sim \operatorname{Multinomial}(m; f_n)$. As long as $m = m(n) \to \infty$ and $m/n \to 0$, we have \vspace{-0.3em}
\[
    \sup _{g\in \operatorname{BL}_1 (\mathbb{R})}\left|\begin{array}{c}{\bbE^{*}\left[ g\left(\sqrt{m}\left\{\psi(f_{n,m}^*) - \psi(f_n) \right\}\right) | f_n\right]} \\ {-\bbE\left[g\left(\sqrt{n}\left\{\psi(f_n) - \psi(f_\star) \right\} \right) \right]}\end{array}\right| \stackrel{p}{\rightarrow} 0.\vspace{-0.3em}
\]
\end{theorem}

\begin{theorem}[Consistency of numerical derivative method]\label{thm:bootConsistency2}
Let $z_n^* \sim \mathcal{N}(\boldsymbol{0}_K, \Sigma(f_n); \mathbb{T})$, a Gaussian distribution truncated in $\mathbb{T}$, where $\mathbb{T} = \mathbb{T}(f_n,\epsilon) = \{x\in \mathbb{R}^K: f_n + \epsilon x \in \mathbb{R}_{+}^K\}$.  As long as $\epsilon = \epsilon(n) \to 0$ and $\sqrt{n}\epsilon \to \infty$, we have \vspace{-0.3em}
\[
    \sup _{g\in \operatorname{BL}_1 (\mathbb{R})}\left|\begin{array}{c}{\bbE^{*}\left[ g\left(\epsilon^{-1}\left\{\psi(f_n + \epsilon z_n^*) - \psi(f_n) \right\}\right) | f_n\right]} \\ {-\bbE\left[g\left(\sqrt{n}\left\{\psi(f_n) - \psi(f_\star) \right\} \right) \right]}\end{array}\right| \stackrel{p}{\rightarrow} 0.\vspace{-0.3em}
\]
\end{theorem}


\subsection{Inference for the audit value}

The preceding bootstrap methods complete our suite of inferential tools for the audit value. In this subsection, we demonstrate the utility of the tools by forming confidence intervals and testing restrictions on the audit value.

One of the most basic inferential tasks is forming a confidence interval of the audit value. Such confidence intervals may be used to give an \emph{asymptotically exact certificate} of individual fairness for ML models. Let $c_q^*$ be the $q$-th quantile of the bootstrap distribution:\vspace{-0.3em}
\[
     c^*_{q} = \inf \{c\in \mathbb{R}: \mathbb{P}(\sqrt{m}\{\psi(f_{n,m}^*) - \psi(f_n)\} \leq c) \geq q \},\vspace{-0.3em}
\]
where $0 \leq q \leq 1$. In practice, $c_q^*$ is estimated by $q$-th quantile of output $\mathcal{S}$ of Algorithm \ref{alg:m-out-of-n} in the Supplementary Materials. Since the approximation error can be made arbitrarily small by increasing number of bootstrap iterations $B$, we ignore this error in our results. 

The two-sided equal-tailed confidence interval for the audit value $\psi(f_\star)$ with asymptotic coverage probability $1-\alpha$ is \vspace{-0.3em}
\begin{equation}\textstyle
     \operatorname{CI_{two-sided}} = \left[\psi(f_n) - \frac{c_{1-\alpha/2}^*}{\sqrt{n}}, \psi(f_n) - \frac{c_{\alpha/2}^*}{\sqrt{n}} \right].
     \label{eq:twoSidedCI}\vspace{-0.3em}
\end{equation}

\begin{theorem}[Asymptotic coverage of two-sided confidence interval]\label{thm:CITwoSide}
For any $f_\star \in \Delta_K$, we have \vspace{-0.3em}
\begin{equation*}
    \liminf_{n \to \infty}  \mathbb{P} \left(\psi(f_\star) \in \operatorname{CI_{two-sided}} \right) \geq 1 - \alpha.\vspace{-0.3em}
\end{equation*}
\end{theorem}

Compared to other certificates of individual fairness (\eg, the certificate in \cite{yurochkin2020Training}), our certificate is asymptotically exact. This is a consequence of the asymptotic exactness of the coverage of the confidence interval \eqref{eq:twoSidedCI}.


Another basic inferential task is testing restrictions on the audit value. In light of the (asymptotic) validity two-sided confidence region \eqref{eq:twoSidedCI}, it is possible to test simple restrictions of the form $\psi(f_*) = \delta$, for some $\delta > 0$, by checking whether $\delta$ falls in  the $(1-\alpha)$-level confidence region. By the duality between confidence intervals and hypothesis tests, this test has asymptotic Type I error rate at most $\alpha$. In the rest of this subsection, we consider the task of testing a compound hypothesis of the form $\psi(f_*) < \delta$.
\begin{definition}
\textbf{\emph{($\delta$--fairness).}} For a constant $\delta\geq0$, an ML system is called $\delta$--fair if $\psi(f_\star) \leq \delta$.
\end{definition}

In order to test whether or not an ML system is $\delta$--fair, the auditor considers hypothesis testing problem \vspace{-0.3em}
\begin{equation}\label{eq:testProblem}
    H_0: \psi(f_\star) \leq \delta~~~~\text{versus}~~~~H_1: \psi(f_\star) > \delta.\vspace{-0.3em}
\end{equation}

The one-sided confidence interval for the audit value $\psi(f_\star)$ with asymptotic coverage probability $1-\alpha$ is \vspace{-0.3em}
\begin{equation*}\textstyle
    \operatorname{CI_{one-sided}} =  \left[\psi(f_n) - \frac{c_{1-\alpha}^*}{\sqrt{n}}, \infty \right).\vspace{-0.3em}
\end{equation*}

We reject the null hypothesis $H_0$ if the one-sided confidence interval does not cover $\delta$, \ie, \vspace{-0.3em}
\begin{equation*}\textstyle
    \delta {\not\in} \left[\psi(f_n) - \frac{c_{1-\alpha}^*}{\sqrt{n}}, \infty \right).\vspace{-0.3em}
\end{equation*}

\begin{theorem}[Asymptotic validity of test]\label{thm:CIOneSide}
For any $\delta \geq 0$, we have \vspace{-0.3em}
\begin{equation*}
    \limsup_{n \to \infty} \sup_{f_{\star} \in \Delta_K: \psi(f_\star) \leq \delta} \mathbb{P}_{f_\star} \left(\delta {\not\in} \operatorname{CI_{one-sided}} \right) \leq \alpha.\vspace{-0.3em}
\end{equation*}
If $\psi(f_\star) > \delta$, then $ \lim_{n \to \infty} \mathbb{P} \left(\delta {\not\in} \operatorname{CI_{one-sided}} \right) = 1$.
\end{theorem}

The choice of threshold $\delta$ is application dependent, and there is no generic recipe to pick $\delta$. It reflects the auditor's tolerance on fairness level of an ML system. For example, in recidivism prediction, a reasonable threshold may be the rate of miscarriage of justice. In other words, the auditor expects the performance of the recidivism prediction instrument to deteriorate by no more than the inherent error rate in the criminal justice system. We demonstrate the suitability of this choice in our computational results.


\begin{figure*}
    \centering
    \includegraphics[width=\textwidth]{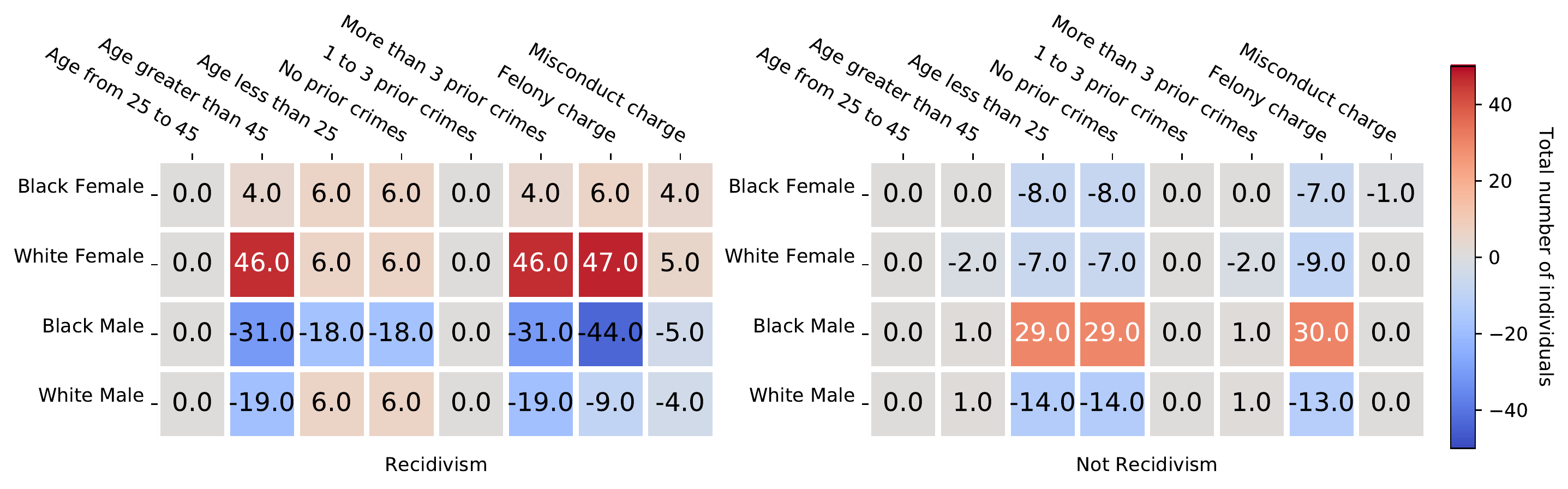}
    \caption{Transport map of vanilla logistic regression on audit dataset. (Number in each grid shows the change in total number of individuals after transport.)}
    \label{fig:transport_compas}
\end{figure*}

\begin{table*}[]
\caption{Numerical comparisons of multiple fairness methods.}
\centering
\begin{tabular}{c|cccccccc}
\hline \hline
{} & FaiTH & CI$^{(2)}_{\textrm{lower}}$ & CI$^{(2)}_{\textrm{upper}}$ & CI$^{(1)}_{\textrm{lower}}$ &  Accuracy & AOD & EOD & SPD \\ \hline
LR & $.06\pm.02$ & $.05\pm.02$ & $.07\pm.03$ & $.05\pm.02$ & $\mathbf{.67}\pm.01$ & $-.23\pm.04$ & $-.19\pm.04$ & $-.26\pm.03$ \\
ADB & $.18\pm.06$ & $.16\pm.05$ & $.20\pm.06$ & $.16\pm.05$ & $.65\pm.01$ & $-.05\pm.13$ & $-.01\pm.12$ & $-.08\pm.13$ \\
RWT & $.15\pm.02$ & $.13\pm.02$ & $.17\pm.02$ & $.14\pm.02$ & $.66\pm.01$ & $\mathbf{-.02}\pm.04$ & $\mathbf{.01}\pm.04$ & $\mathbf{-.06}\pm.04$ \\
LFR & $.07\pm.05$ & $.06\pm.04$ & $.08\pm.05$ & $.06\pm.05$ & $.66\pm.01$ & $-.09\pm.09$ & $-.06\pm.07$ & $-.13\pm.08$ \\
RLR & $\mathbf{.02}\pm.02$ & $\mathbf{.01}\pm.02$ & $\mathbf{.02}\pm.02$ & $\mathbf{.01}\pm.02$ & $.66\pm.01$ & $-.19\pm.03$ & $-.15\pm.03$ &               $-.22\pm.03$ \\
\hline \hline
\end{tabular}
\label{tab:my_label}
\end{table*}

\section{Computational results}
\label{sec:computationalResults}

We shall verify correctness of our methodology using widely studied COMPAS dataset \citep{angwin2016Machine}. Originally it was shown that COMPAS score used for providing recommendation to the judge if a person will recommit or not is biased against certain groups of individuals. In \cite{angwin2016Machine}, it was shown that COMPAS score is strongly biased against men and minorities.


To apply our methodology it remains to choose metric and loss function for the auditor's problem. We make choices to facilitate simplicity and interpretability of the analysis. For the metric we consider any two observations which only differ in race or gender to have distance zero between each other and infinity otherwise. For the loss we shall consider 0-1 loss, then FaiTH value can be understood as missclassification rates induced by the solution of the auditor's problem \eqref{eq:auditorsProblem} and threshold $\delta$ corresponds to the amount of classification errors that the auditor believes it is justified for the problem. Here we choose $\delta = 0.0365$, which is the midpoint of the results reported by various studies on the number of innocent prisoners in the United States \citep{wiki_justice}.

\subsection{Audit guidelines and interpretation}
In this subsection we give practical guidelines for an auditor wishing to assess performance of an ML system. We will investigate performance of a vanilla logistic regression (LR) classifier trained on COMPAS dataset to predict if a person will re-offend. We use 70\% of the COMPAS dataset to train the classifier and the remaining 30\% to audit it using black-box access to the trained model. To determine if an ML system is individually fair we compute the FaiTH value and report lower and upper bounds of the 95\% two-sided confidence interval (CI$^{(2)}_{\textrm{lower}}$ and CI$^{(2)}_{\textrm{upper}}$) and lower bound of the 95\% one-sided confidence interval (CI$^{(1)}_{\textrm{lower}}$) using methodology described in the preceding sections. We fail to reject the hypothesis that a classifier is individually fair if a pre-specified value of $\delta$ is contained in the confidence interval. 


We repeat the experiment 50 times and summarize the results in Table \ref{tab:my_label}. Common group-fairness metrics are reported and FaiTH is applied to test previously proposed fair classification techniques motivated by the notion of group fairness. Before discussing the relation to group fairness, we complete the audit analysis of the logistic regression. Both one- and two-sided confidence intervals lower bounds are equal to $0.05 > \delta$ on average, meaning that auditor should reject the individual fairness hypothesis of the logistic regression classifier.

In this situation auditor may utilize the adversarial distribution computed to evaluate the FaiTH statistics in \eqref{eq:auditorsProblem} to investigate the patterns of individual fairness violation. We present such analysis in Figure \ref{fig:transport_compas}. On the left heat map we show the change in distribution of the features of individuals labeled as recidivists in the audit data (counts of the distribution maximizing \eqref{eq:auditorsProblem} minus counts of the audit dataset distribution). We can interpret the figure column-wise: there are 31 black males and 19 white males older than 45 that were correctly classified as recidivists, but would be misclassified as non-reoffenders if they were to be white females (or black females for the 4 of them); similar argument holds for recidivists with more than 3 prior crimes and/or a felony charge. In summary, we see that white females are treated by the classifier as a privileged group. The right figure shows analogous heat map for individuals labeled as non-reoffenders in the audit data. Among others we see that young white males and females, and black females correctly classified to not commit recidivism would be classified as recidivists if they were to be black males. Previous study of the COMPAS dataset reports white females as the privileged group and black males as unprivileged \citep{propublica_compas}, aligning with our findings. We can also make an additional observation based on our analysis: people in the age group of 25 to 45 and/or those with 1 to 3 prior crimes were treated individually fair by the classifier. Auditor may utilize such findings to provide recommendations to the ML system provider if the system fails to pass the FaiTH test without disclosing the audit data.

\paragraph{Relation to group fairness} We proceed to evaluate the individual fairness hypothesis for several group fairness approaches proposed in the literature. We consider three algorithms available in the IBM AIF360 toolkit \citep{aif360-oct-2018}. Two pre-processing techniques: Reweighting (RWT) \citep{kamiran2012data} that modifies data weights in the training loss, and Learning Fair Representation (LFR) \citep{zemel2013Learning} that finds transformed feature space obfuscating information about protected attributes. And an in-processing technique: Adversarial Debiasing (ADB) \citep{zhang2018mitigatingSK} that learns a group-fair predictor by reducing the ability of a corresponding adversary to predict protected attributes. We also report common group fairness metrics (for all prefered value is close to 0): average odds difference (AOD), equal opportunity difference (EOD) and statistical parity difference (SPD). Results are summarized in Table \ref{tab:my_label}: all of these methods succeed in reducing the group biases, however they tend to exacerbate individual fairness violations as can be seen from the FaiTH value. For example, Reweighting method appears to mitigate most of the group biases, but investigating corresponding logistic regression fit we find that it assigns large coefficient to the race variable. In other words, decision of the corresponding classifier is majorly affected by the race, which is not permissible from the perspective of individual fairness and an alarm is raised by FaiTH.


\subsection{Model selection under FaiTH constraint}
\label{sec:faith_reg}

In this subsection, we propose a generic model selection strategy under $\delta$-fairness constraint, and present the strategy by logistic regression with $\ell_1$ penalty.

The idea of strategy is to select candidates of models which pass the fairness hypothesis testing \eqref{eq:testProblem}. To be precise, we filter all models through comparison between the fairness threshold $\delta$ and the CI lower bound of audit value evaluated on validation dataset. Then among these candidates, we select the model which has the lowest validation error.

The dataset is splited into training, validation, and audit dataset. We fit $\ell_1$-regularized logistic regression (RLR) by minimizing $\mathcal{L}(Z, \beta) + \lambda \|\beta\|_1$,
where $\beta$ is vector of regression coefficients, $Z$ is the training set, $\mathcal{L}$ is the logistic loss, and $\lambda > 0$ is a tuning parameter.

\begin{figure}[htbp]
    \centering
    \vspace{-0.115in}
    \includegraphics[width=0.49\textwidth]{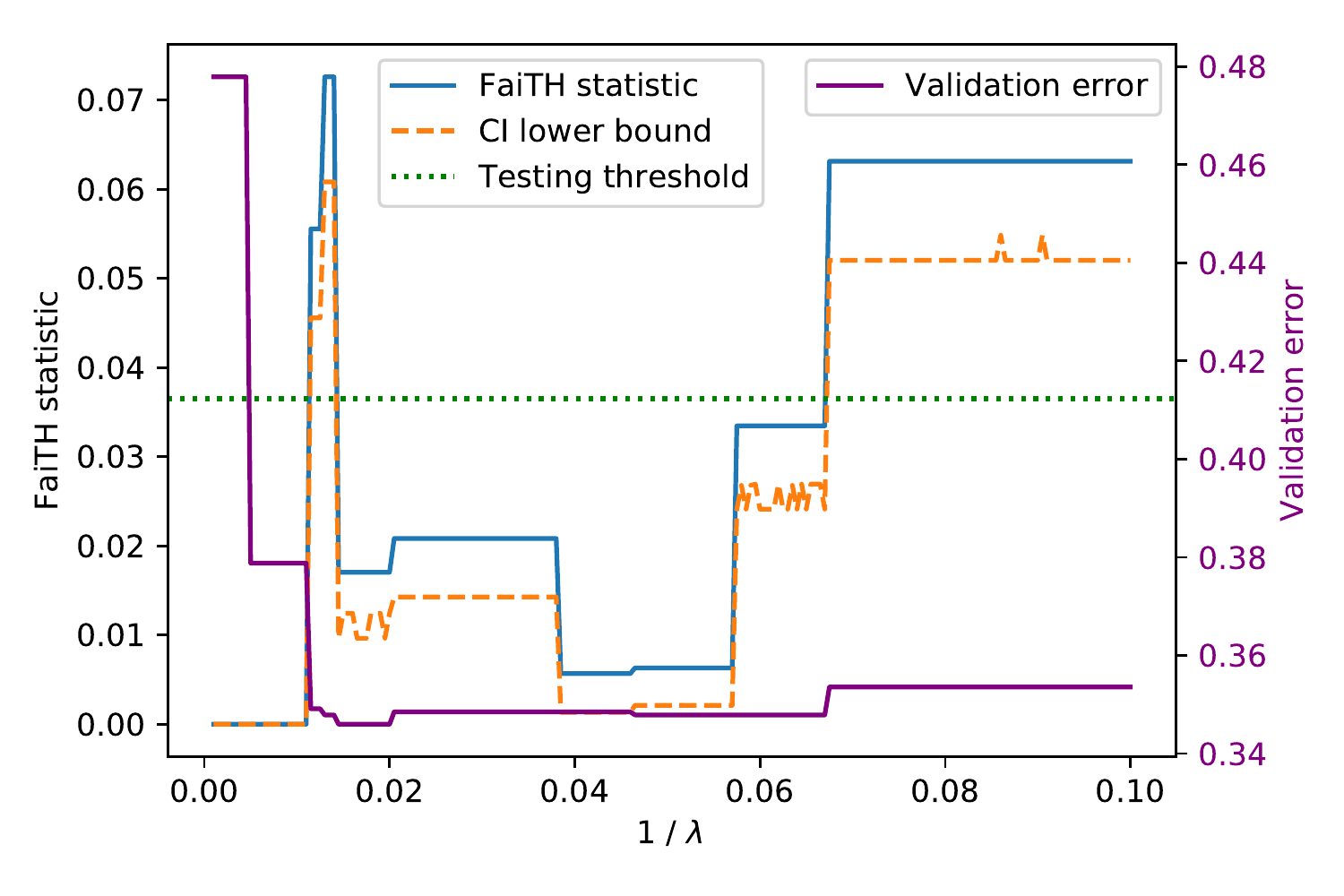}
    \vspace{-0.4in}
    \caption{Performance of logistic regression with $\ell_1$ penalty on validation dataset.}
    \label{fig:regularization}
    \vspace{-0.15in}
\end{figure}

Figure \ref{fig:regularization} demonstrates trade-off between accuracy and fairness. Strong penalty (\ie, small value of $\frac1\lambda$) results in tiny FaiTH statistic but huge validation error, and on the contrary, weak penalty (\ie, large value of $\frac1\lambda$) leads to undesirable fairness level but satisfactory accuracy. The broken orange line shows lower bounds of 95\% confidence interval (one-sided) of validation audit value for each $\lambda$. Note that a tuning parameter $\lambda$ passes the $\delta$--fairness test if and only if its corresponding CI lower bound is smaller than $\delta$, so the range of that orange broken line lies under green dotted line determines all candidates of $\delta$--fair tuning parameters. Choosing the tuning parameter which has lowest validation error among these candidates outputs the selected $\frac1\lambda = 0.0145$. We note that gender is not selected so that prediction without using gender can effectively ensure model's individual fairness and keep comparable prediction accuracy at the same time. 

\begin{figure}[htbp]
    \centering
    \includegraphics[width=0.49\textwidth]{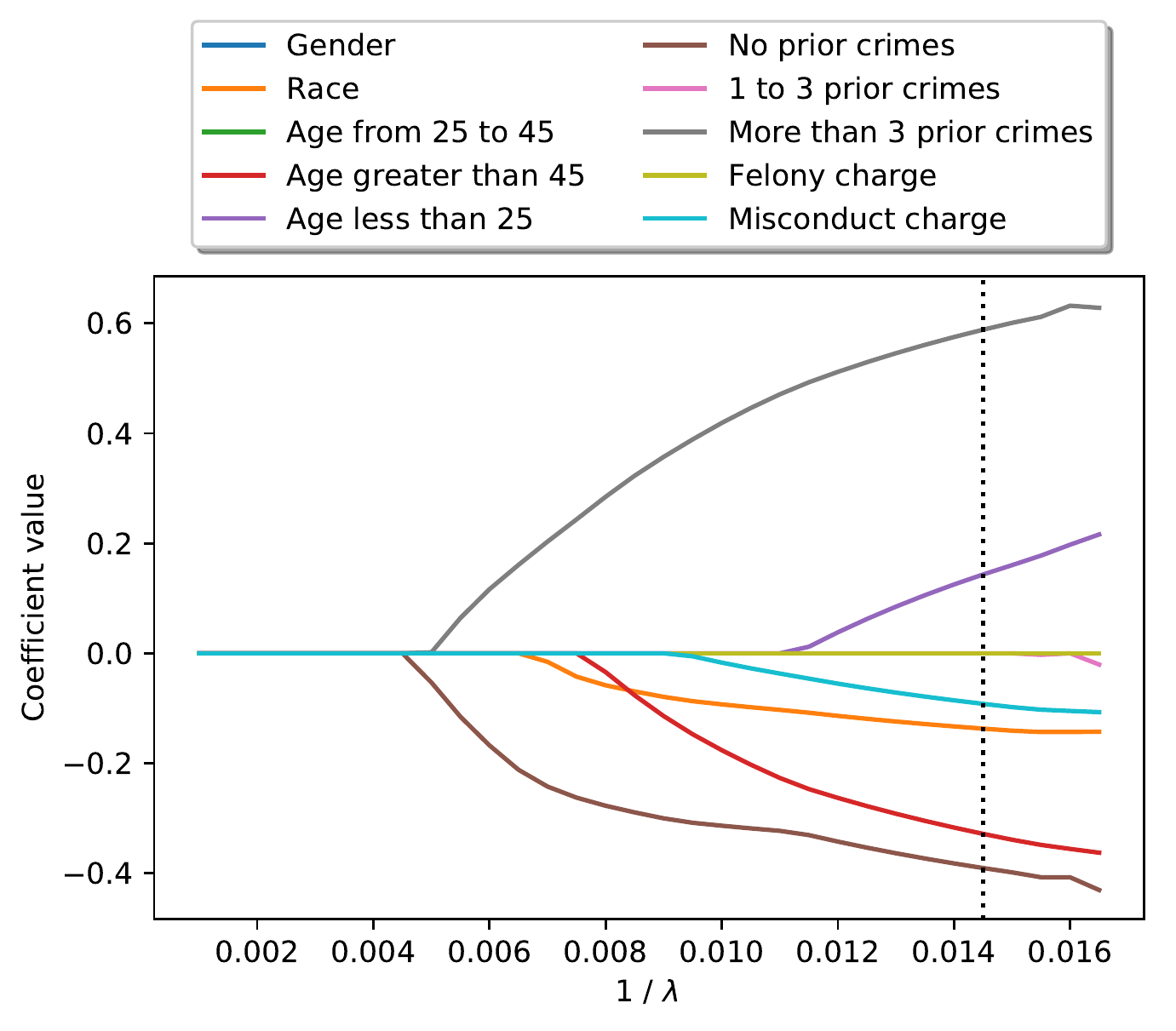}
    \vspace{-0.35in}
    \caption{Solution pathes of logistic regression with $\ell_1$ penalty.}
    \label{fig:solution-path}
\end{figure}

Solution pathes of regression coefficients are depicted in Figure \ref{fig:solution-path}. The vertical dotted line $\frac1\lambda = 0.0145$ shows the selected model. Whether or not an individual has prior crimes is of the greatest significance for predicting recidivism since the corresponding coefficient pops out firstly. The other five selected variables are ``more than 3 prior crimes'', race, ``age greater than 45'', ``misconduct charge'', and ``age less than 25'' in sequence. 

We run our model selection strategy for 50 times and make comparison with other methods in Table \ref{tab:my_label}. RLR continues to have low FaiTH value when we computed on the audit dataset and is the only method for which we fail to reject the individual fairness hypothesis. RLR also has better group fairness scores than the baseline, however not as good as those of other group fairness approaches. We note that RLR is a simple model selection based approach that is plausible due to the development of our FaiTH methodology. Combining FaiTH with prior ideas used for group fairness may layout a pass for training ML systems with strong guarantees for both individual and group fairness.

\section{Summary and discussion}
\label{sec:summary}

In this paper, we developed a suite of inferential tools for detecting and localizing individual bias/unfairness in the ML model. Our tools only require black-box access to the ML model and are computationally efficient. Further, they allow auditors to control the false alarm rate and provide asymptotically exact certificates of fairness. We demonstrated the utility of our tools by using them to reveal the gender and racial biases in Northpointe's COMPAS recidivism prediction instrument. 

\newpage
\subsubsection*{Acknowledgements}

This work was supported by the National Science Foundation under grants DMS-1830247 and DMS-1916271.

\bibliographystyle{plainnat} 
\bibliography{YK, MY, SK}

\onecolumn
\appendix

{\centering \Large \textbf{Supplementary Materials for}\\}
{\centering \Large \textbf{``Auditing ML Models for Individual Bias and Unfairness''}\\}
\section{Proofs}
\label{sec:proofs}

\subsection{Proof of Proposition in Section 2}
\emph{Proof of Proposition \ref{prop:robustFaiTH}.} For the simplicity of notations, we drop the subscript of the loss function picked by the auditor, that is, we denote $\ell_h$ by $\ell$. Furthermore, let
\begin{equation*}
    \ell^c_{\lambda}(z) = \ell^c_{\lambda}(x, y) \triangleq \sup_{x_2 \in \cX}\left\{\ell(x_2, y) - \lambda c((x,y), (x_2, y))\right\}.
\end{equation*}
By the duality result of \cite{blanchet2019quantifying}, for any $\varepsilon > 0$, we have
\begin{equation*}
    \sup_{P: W(P, P_n) \leq \varepsilon} \bbE_{Z \sim P}[\ell(Z)] = \inf_{\lambda \geq 0}\left\{\lambda\varepsilon + \bbE_{Z \sim P_n}[\ell^c_{\lambda}(Z)]\right\}
\end{equation*}
and
\begin{equation*}
    \sup_{P: W_*(P, P_n) \leq \varepsilon} \bbE_{Z \sim P}[\ell(Z)] = \inf_{\lambda \geq 0}\left\{\lambda\varepsilon + \bbE_{Z \sim P_n}[\ell^{c_*}_{\lambda}(Z)]\right\}.
\end{equation*}
Let $\lambda_* \in \arg \min_{\lambda\geq 0} \left\{\lambda\varepsilon + \bbE_{Z \sim P_n}[\ell^{c_*}_{\lambda}(Z)]\right\}$. Then we have
\begin{align*}
    &\sup_{P: W(P, P_n) \leq \varepsilon} \bbE_{Z \sim P}[\ell(Z)] - \sup_{P: W_*(P, P_n) \leq \varepsilon} \bbE_{Z \sim P}[\ell(Z)] \\
    &= \inf_{\lambda \geq 0}\left\{\lambda\varepsilon + \bbE_{Z \sim P_n}[\ell^c_{\lambda}(Z)]\right\} - \lambda_*\varepsilon - \bbE_{Z \sim P_n}[\ell^{c_*}_{\lambda_*}(Z)] \\
    &\leq \lambda_*\varepsilon + \bbE_{Z \sim P_n}[\ell^{c}_{\lambda_*}(Z)] - \lambda_*\varepsilon - \bbE_{Z \sim P_n}[\ell^{c_*}_{\lambda_*}(Z)] \\
    &= \bbE_{Z \sim P_n}[\ell^{c}_{\lambda_*}(Z) - \ell^{c_*}_{\lambda_*}(Z)].
\end{align*}
By Assumption A3, we have
\begin{align*}
    \ell^{c}_{\lambda_*}(z) - \ell^{c_*}_{\lambda_*}(z) & = \sup_{x_2 \in \cX}\left\{\ell(x_2, y) - \lambda_* c((x,y), (x_2, y))\right\} - \sup_{x_2 \in \cX}\left\{\ell(x_2, y) - \lambda_* c_*((x,y), (x_2, y))\right\} \\
    &\leq \lambda_* \sup_{x_2\in \cX}\left|c((x,y), (x_2, y)) - c_*((x,y), (x_2, y)) \right| \\
    &\leq \lambda_* \eta D^2.
\end{align*}
Thus, we conclude that
\begin{equation*}
    \sup_{P: W(P, P_n) \leq \varepsilon} \bbE_{Z \sim P}[\ell(Z)] - \sup_{P: W_*(P, P_n) \leq \varepsilon} \bbE_{Z \sim P}[\ell(Z)] \leq \lambda_* \eta D^2.
\end{equation*}
Similarly, we have
\begin{equation*}
    \sup_{P: W_*(P, P_n) \leq \varepsilon} \bbE_{Z \sim P}[\ell(Z)] - \sup_{P: W(P, P_n) \leq \varepsilon} \bbE_{Z \sim P}[\ell(Z)] \leq \lambda_\dagger \eta D^2,
\end{equation*}
where $\lambda_\dagger \in \arg \min_{\lambda\geq 0} \left\{\lambda\varepsilon + \bbE_{Z \sim P_n}[\ell^{c}_{\lambda}(Z)]\right\}$.

\vspace{1em}
Now, it suffices to show that $\lambda_* \leq \frac{L}{\sqrt{\varepsilon}}$ (and similarly $\lambda_\dagger \leq \frac{L}{\sqrt{\varepsilon}}$). By the optimality of $\lambda_*$,
\begin{align*}
    \lambda_* \varepsilon &\leq \lambda_* \varepsilon + \bbE_{Z\sim P_n}[\sup_{x_2\in\cX}\{\ell(x_2, Y) - \lambda_* d_{x_*}^2(X, x_2)\} - \ell(X,Y)] \\
    &= \lambda_* \varepsilon + \bbE_{Z\sim P_n}[\ell^{c_*}_{\lambda_*}(Z) - \ell(Z)] \\
    &\leq \lambda\varepsilon + \bbE_{Z\sim P_n}[\ell^{c_*}_{\lambda}(Z) - \ell(Z)] \\
    &= \lambda \varepsilon + \bbE_{Z\sim P_n}[\sup_{x_2\in\cX}\{\ell(x_2, Y) - \ell(X,Y) - \lambda d_{x_*}^2(X, x_2)\}]
\end{align*}
for any $\lambda \geq 0$. By Assumption A2, the right-hand side is at most
\begin{align*}
    \lambda_* \varepsilon &\leq \lambda \varepsilon + \bbE_{Z\sim P_n}[\sup_{x_2\in\cX}\{L d_{x_*}(X, x_2) - \lambda d_{x_*}^2(X, x_2)\}] \\
    &\leq \lambda\varepsilon + \sup_{t\geq 0}\{Lt - \lambda t^2\}.
\end{align*}
We minimize the right-hand side with respect to $t$ (set $t = \frac{L}{2\lambda}$) and $\lambda$ (set $\lambda = \frac{L}{2\sqrt{\varepsilon}}$) to obtain $\lambda_* \varepsilon \leq L\sqrt{\varepsilon}$, or equivalently $\lambda_* \leq \frac{L}{\sqrt{\varepsilon}}$. \hfill $\square$

\subsection{Proofs of Theorems in Section 3}
\emph{Proof of Theorem \ref{thm:asymptoticDistribution}.} We are working with Euclidean space $\mathbb{D} = \mathbb{R}^{K}$ and $\mathbb{E} = \mathbb{R}$.

By Theorem \ref{thm:HadamardDerivative}, $\psi: \mathbb{R}^{K} \to \mathbb{R}$ is Hadamard directionally differentiable at $f_\star$ (tangentially to $\mathbb{R}^{K}$).

Since $f_n$ is the empirical version of $f_\star$, by central limit theorem, we have
\begin{equation*}
\sqrt{n}(f_{n}-f_{\star}) \stackrel{d}{\rightarrow} \mathcal{N}(0, \Sigma(f_\star)) \stackrel{d}{\sim} Z,
\end{equation*}

which is tight and supported in $\mathbb{R}^{K}$.

Via delta method (Theorem \ref{thm:deltaMethod}) with $\psi(\cdot)$ and the derivative formula given by Theorem \ref{thm:HadamardDerivative}, we conclude
\begin{equation*}
    \sqrt{n}\{\psi(f_n) - \psi(f_\star)\} \stackrel{d}{\rightarrow} \psi_{f_\star}^\prime(Z)= \inf \{ (\lambda + l)^\top Z: (\nu, \mu, \lambda) \in \Lambda \}.
\end{equation*}

Hence we complete the proof of Theorem \ref{thm:asymptoticDistribution}. \hfill $\square$

\vspace{1em}
The next theorem adapted from from \cite{bonnans2000Perturbation} will turn out to be useful.

\begin{theorem}[Proposition 4.27 in \cite{bonnans2000Perturbation}]\label{thm:BonnansShapiro}
$\mathbb{A}$, $\mathbb{B}$ and $\mathbb{V}$ are Banach spaces. $f: \mathbb{A} \to \mathbb{R}$ is continuously differentiable. $G + \bullet: \mathbb{A} \times \mathbb{V} \to \mathbb{B}$ is continuously differentiable. $\mathbb{K}$ is a closed convex subset of $\mathbb{B}$. Consider a class of problems 
$$
\begin{aligned}
&(\mathcal{P}_v):~~~~& & \min_{x \in \mathbb{A}} & & f(x) \\
& & & \operatorname{subject~to} & & G(x) + v \in \mathbb{K}
\end{aligned}
$$
parameterized by $v \in \mathbb{V}$. Let $\varphi(v)$ be the optimal value of the problem $\mathcal{P}_v$. Suppose that
\begin{enumerate}
    \item for $v = 0$, the problem $\mathcal{P}_0$ is convex;
    \item $\varphi(0)$ is finite;
    \item $0 \in \operatorname{int}\{G(\mathbb{A}) - \mathbb{K}\}$.
\end{enumerate}
Then the optimal value function $\varphi(v)$ is Hadamard directionally differentiable at $v = 0$. Furthermore, 
\[
    \lim_{h^\prime \to h, t \to 0^+} \frac{\varphi(th^\prime) - \varphi(0)}{t} = \sup \{\lambda^\top h: \lambda \in \Gamma\}
\]
for any $h \in \mathbb{V}$, where $\Gamma$ is the set of optimal solutions of the dual problem of $\mathcal{P}_0$.
\end{theorem}

\vspace{1em}
\emph{Proof of Theorem \ref{thm:HadamardDerivative}.} We first prove the theorem without constraint $\langle D, \Pi \rangle = 0$. In order to employ Theorem \ref{thm:BonnansShapiro}, the result of canonical perturbation, we introduce a parameter $t\in \mathbb{R}$, and the optimization problem $\psi(f_\star)$ can be equivalently rewritten as

$$
\begin{aligned}
&(\text{P1}):~~~~& & \max_{t\in \mathbb{R}, \Pi\in \mathbb{R}_{+}^{K\times K}} & &l^\top (\Pi^\top \boldsymbol{1}_K - f_\star) + t & & \\
& & & \operatorname{subject~to} & & \langle C,\Pi\rangle \leq \varepsilon & & :\nu\\
& & & & &\textstyle\Pi \boldsymbol{1}_K = f_\star & & :\lambda\\
& & & & & t = 0 & & :\eta
\end{aligned}
$$

where $\nu, \lambda, \eta$ are Lagrange multipliers. 

The canonical perturbation of problem (P1) is then given by 

$$
\begin{aligned}
&(\mathcal{P}_{u,v,w}):~~~~& & \max_{t\in \mathbb{R}, \Pi\in \mathbb{R}_{+}^{K\times K}} & &l^\top (\Pi^\top \boldsymbol{1}_K - f_\star) + t \\
& & & \operatorname{subject~to} & & \langle C,\Pi\rangle + u \leq \varepsilon \\
& & & & &\textstyle\Pi \boldsymbol{1}_K + v = f_\star \\
& & & & & t + w = 0,
\end{aligned}
$$

which outputs its optimal value $\varphi(u, v, w)$. Thus $\varphi$ is a function from $\mathbb{R}^{K+2}$ to $\mathbb{R}$.

Let $\mathbb{A} = \mathbb{R}_{+}^{K\times K} \times \mathbb{R}$, $\mathbb{B} = \mathbb{V} = \mathbb{R}^{K+2}$, and $\mathbb{K} = \{(x,f_\star^\top,0)^\top: x \leq \varepsilon \} \subset \mathbb{R}^{K+2}$. Consider function $f:\mathbb{A} \to \mathbb{R}$ such that $(\Pi, t) \mapsto -\{ l^\top (\Pi^\top \boldsymbol{1}_K - f_\star) + t \}$, and function $G:\mathbb{A} \to \mathbb{B}$ such that $(\Pi, t) \mapsto (\langle C,\Pi\rangle, (\Pi \boldsymbol{1}_K)^\top, t)^\top$.

Then, the class of maximization problems $(\mathcal{P}_{u,v,w})$ is equivalent to the following class of minimization problems 
$$
\begin{aligned}
&(\mathcal{Q}_{u,v,w}):~~~~& & \min_{(\Pi, t) \in \mathbb{A}} & & f(\Pi, t) \\
& & & \operatorname{subject~to} & & G(\Pi, t) + (u,v^\top, w)^\top \in \mathbb{K}.
\end{aligned}
$$

Denote the optimal value function of $\mathcal{Q}_{u,v,w}$ by $\phi(u,v,w)$.

(i) To check item 1 in Theorem \ref{thm:BonnansShapiro}, we note that $\mathcal{Q}_{0, \boldsymbol{0}_{K}, 0}$ is a problem of linear programming, and thus a convex optimization problem.

(ii) Item 2 in Theorem \ref{thm:BonnansShapiro} is guaranteed by
\begin{equation*}
    \varepsilon \geq 0 = \min \{\langle C, \Pi \rangle: \Pi\in \mathbb{R}_{+}^{K\times K}, \Pi \boldsymbol{1}_K = f_\star\},
\end{equation*}

which implies that $\mathcal{Q}_{0, \boldsymbol{0}_{K},0}$ has a solution, and thus $\phi(0, \boldsymbol{0}_{K}, 0)$ is finite.

(iii) $f_\star \in \mathbb{R}_{+}^K$ ensures that item 3 in Theorem \ref{thm:BonnansShapiro} holds.

Now applying Theorem \ref{thm:BonnansShapiro} to $(\mathcal{Q}_{u,v,w})$, we conclude that $\phi$ is Hadamard directionally differentiable at the origin. Note that $\varphi = -\phi$, we can further conclude that $\varphi$ is also Hadamard directionally differentiable at the origin, and

\begin{equation*}
    \lim_{\substack{\xi^\prime \to \xi \\ t \to 0^+}} \frac{\varphi(0, t\xi^\prime) - \varphi(0, \boldsymbol{0}_{K+1})}{t} = -\lim_{\substack{\xi^\prime \to \xi \\ t \to 0^+}} \frac{\phi(0, t\xi^\prime) - \phi(0, \boldsymbol{0}_{K+1})}{t} = - \sup \{ \langle(\lambda^\top, w)^\top, \xi\rangle: (\nu, \lambda, w) \in \Gamma \},
\end{equation*}

where $\Gamma$ is the set of optimal solutions of the dual problem of (P1).

Furthermore, one can check that $\Gamma = \Lambda \times \{-1\}$, where $\Lambda$ is the set of optimal solutions of the dual problem of $\psi(f_\star)$.

Specifically, the dual problem of $\psi(f_\star)$ is given by
$$
\begin{aligned}
& \min_{\nu\geq 0, \lambda_1, \cdots, \lambda_K} & &-\varepsilon \nu - \sum_{k=1}^K f_\star^{(k)} \lambda_k \\
& \operatorname{subject~to} & & c_{ij}\nu + \lambda_i \leq -l_j,~~\text{for}~1\leq i, j \leq K.
\end{aligned}
$$

Thus, we have
\begin{equation*}
    \Lambda = \underset{\nu, \geq 0, \lambda \in \mathbb{R}^K}{\argmax} \{ \varepsilon \nu + f_\star^\top \lambda: c_{ij}\nu + \lambda_i \leq -l_j, 1\leq i,j \leq K\}
\end{equation*}

Note that $\psi(f) = \varphi(0, f_\star - f, l^\top (f-f_\star))$, we conclude that $\psi(f)$ is Hadamard directionally differentiable at $f_\star$, and the derivative formula is given by
\begin{align*}
    \psi_{f_\star}^\prime(h) &= \lim_{\substack{h^\prime \to h \\ t \to 0^+}} \frac{\psi(f_\star + th^\prime) - \psi(f_\star)}{t}\\
    &= \lim_{\substack{h^\prime \to h \\ t \to 0^+}} \frac{\varphi(0, -th^\prime, tl^\top h^\prime) - \varphi(0, \boldsymbol{0}_K, 0)}{t} \\
    &= \lim_{\substack{\xi^\prime \to \xi \\ t \to 0^+}} \frac{\varphi(0, t\xi^\prime) - \varphi(0, \boldsymbol{0}_{K+1})}{t}~~~~\left[\text{where}~\xi = (-h^\top, l^\top h)^\top\right] \\
    &= - \sup \{ \langle(\lambda^\top, w)^\top, \xi\rangle: (\nu, \lambda, w) \in \Gamma \}\\
    &= - \sup \{ \langle(\lambda^\top, -1)^\top, (-h^\top, l^\top h)^\top\rangle: (\nu, \lambda) \in \Lambda \}\\
    &= -\sup \{-\langle \lambda + l, h\rangle: (\nu, \lambda) \in \Lambda\}\\
    &= \inf \{\langle \lambda + l, h\rangle: (\nu, \lambda) \in \Lambda\}.
\end{align*}

For the case with constraint $\langle D, \Pi \rangle = 0$, note that the dual problem of $\psi(f_\star)$ changes slightly into
$$
\begin{aligned}
& \min_{\nu, 
\mu\geq 0, \lambda_1, \cdots, \lambda_K} & &-\varepsilon \nu - \sum_{k=1}^K f_\star^{(k)} \lambda_k \\
& \operatorname{subject~to} & & c_{ij}\nu + d_{ij}\mu + \lambda_i \leq -l_j,~~\text{for}~1\leq i, j \leq K,
\end{aligned}
$$
and
\begin{equation*}
    \Lambda = \underset{\nu,\mu \geq 0, \lambda \in \mathbb{R}^K}{\argmax} \{ \varepsilon \nu + f_\star^\top \lambda: c_{ij}\nu + d_{ij}\mu + \lambda_i \leq -l_j, 1\leq i,j \leq K\}.
\end{equation*}

Hence we complete the proof of Theorem \ref{thm:HadamardDerivative}. \hfill $\square$

\subsection{Proofs of Theorems in Section 4}
The following lemma adapted from \cite{hong2018numerical} provides a general recipe for the consistency of our two bootstrap strategies.

\begin{lemma}[Theorem 3.1 in \cite{hong2018numerical}]\label{lemma:lemma}
Suppose $\mathbb{D}$ and $\mathbb{E}$ are Banach Spaces and $\phi : \mathbb{D}_{\phi} \subseteq \mathbb{D} \mapsto \mathbb{E}$ is Hadamard directionally differentiable at $\theta_{0}$ tangentially to $\mathbb{D}_{0}$. Let $\hat{\theta}_{n} :\left\{X_{i}\right\}_{i=1}^{n} \mapsto \mathbb{D}_{\phi}$ be such that for some $r_{n} \uparrow \infty, r_{n}\left\{\hat{\theta}_{n}-\theta_{0}\right\} \leadsto \mathbb{G}_{0}$ in $\mathbb{D}$, where $\mathbb{G}_0$ is tight and its support is included in $\mathbb{D}_{0}$. Then
\begin{equation*}
    r_{n}\left(\phi\left(\hat{\theta}_{n}\right)-\phi\left(\theta_{0}\right)\right) \leadsto \phi_{\theta_{0}}^{\prime}\left(\mathbb{G}_{0}\right).
\end{equation*}
Let $\mathbb{Z}_{n}^{*} \leadsto \mathbb{G}_{0}$ satisfy regularity of measurability \footnote{$\mathbb{Z}_{n}^{*}$ is asymptotically measurable jointly in the data and the bootstrap weights; $g\left(\mathbb{Z}_{n}^{*}\right)$ is a measurable function of the bootstrap weights outer almost surely in the data for
every bounded, continuous map $g : \mathbb{D} \to \mathbb{R}$; $\mathbb{G}_{0}$ is Borel measurable and separable.}. Then for $\epsilon_{n} \rightarrow 0, r_{n} \epsilon_{n} \rightarrow \infty$,
\begin{equation*}
    \hat{\phi}_{n}^{\prime}\left(\mathbb{Z}_{n}^{*}\right) \stackrel{\operatorname{def}}{=\joinrel=} \frac{\phi\left(\hat{\theta}_{n}+\epsilon_{n} \mathbb{Z}_{n}^{*}\right)-\phi\left(\hat{\theta}_{n}\right)}{\epsilon_{n}} \leadsto \phi_{\theta_{0}}^{\prime}\left(\mathbb{G}_{0}\right).
\end{equation*}
\end{lemma}

\vspace{1em}

\emph{Proof of Theorem \ref{thm:bootConsistency1}.} Hereafter, $\mathbb{G}_0$ refers to $\mathcal{N}(f_\star, \Sigma(f_\star))$. By central limit theorem, we have
\begin{equation*}
    \sqrt{n}\{f_n -f_\star\} \leadsto \mathbb{G}_0~~\text{and}~~\sqrt{m}\{f_{n,m}^* - f_\star\} \leadsto \mathbb{G}_0.
\end{equation*}
Since $m/n \to 0$, we have
\begin{equation*}
    \sqrt{m}\{f_{n,m}^* - f_n\} = \sqrt{m}\{f_{n,m}^* - f_\star\} - \sqrt{\frac{m}{n}} \sqrt{n}\{f_n -f_\star\} \leadsto \mathbb{G}_0.
\end{equation*}
Let $r_n = \sqrt{n}, \epsilon_n = 1/\sqrt{m}$ and $\mathbb{Z}_n^\star = \sqrt{m}\{f_{n,m}^* - f_n\}$. Then $\epsilon_{n} \rightarrow 0, r_{n} \epsilon_{n} \rightarrow \infty$, and $\mathbb{Z}_n^\star \leadsto \mathbb{G}_0$. Applying Lemma \ref{lemma:lemma}, we conclude
\begin{align*}
    \sqrt{m} \left\{\psi(f_{n,m}^*) - \psi(f_n) \right\} &= \frac{\psi\left(f_n + \frac{1}{\sqrt{m}}\sqrt{m}\{f_{n,m}^* - f_n\}\right) - \psi(f_n)}{1/\sqrt{m}} \\
    &= \frac{\psi(f_n + \epsilon_n \mathbb{Z}_n^*) - \psi(f_n)}{\epsilon_n} \leadsto \psi^{\prime}_{f_\star}(\mathbb{G}_0).
\end{align*}

Finally, note that $\sqrt{n}\{\psi(f_n)-\psi(f_\star)\} \leadsto \psi_{f_\star}^\prime(\mathbb{G}_0)$, we have
\begin{align*}
    &\sup_{g\in \operatorname{BL}_1 (\mathbb{R})} \Big|\mathbb{E}\left[ g\left(\sqrt{m} \left\{\psi(f_{n,m}^*) - \psi(f_n) \right\}\right) | f_n\right] -  \mathbb{E}\left[g\left(\sqrt{n}\left\{\psi(f_n) - \psi(f_\star) \right\} \right) \right] \Big| \\
    \leq &\sup_{g\in \operatorname{BL}_1 (\mathbb{R})} \Big|\mathbb{E}\left[ g\left(\sqrt{m} \left\{\psi(f_{n,m}^*) - \psi(f_n) \right\}\right) | f_n\right] -  \mathbb{E}\left[g\left(\psi^{\prime}_{f_\star}(\mathbb{G}_0) \right) \right] \Big| \\
    & + \sup_{g\in \operatorname{BL}_1 (\mathbb{R})} \Big| \mathbb{E}\left[g\left(\psi^{\prime}_{f_\star}(\mathbb{G}_0) \right) \right] - \mathbb{E}\left[g\left(\sqrt{n}\left\{\psi(f_n) - \psi(f_\star) \right\} \right) \right] \Big| \\
    &= o_p(1) + o_p(1) = o_p(1)
\end{align*}

by triangle inequality. Hence we complete the proof of Theorem \ref{thm:bootConsistency1}. \hfill $\square$

\vspace{1em}

\emph{Proof of Theorem \ref{thm:bootConsistency2}.} By central limit theorem, we have
\begin{equation*}
    \sqrt{n}\{f_n -f_\star\} \leadsto \mathbb{G}_0 \sim \mathcal{N}(\boldsymbol{0}_k, \Sigma(f_\star)).
\end{equation*}

As $\epsilon \to 0, n \to \infty$, we have
\begin{equation*}
    \mathbb{T}(f_n,\epsilon) \to \mathbb{R}^K~~\text{and}~~z_n^* \sim \mathcal{N}(\boldsymbol{0}_K, \Sigma(f_n); \mathbb{T}) \leadsto \mathcal{N}(\boldsymbol{0}_k, \Sigma(f_\star)) \sim \mathbb{G}_0.
\end{equation*}

Let $r_n = \sqrt{n}, \epsilon_n = \epsilon$, and $\mathbb{Z}_n^* = z_n^*$. Then $\epsilon_{n} \rightarrow 0, r_{n} \epsilon_{n} \rightarrow \infty$, and $\mathbb{Z}_n^\star \leadsto \mathbb{G}_0$. Applying Lemma \ref{lemma:lemma}, we conclude
\begin{equation*}
    \epsilon^{-1} \left\{\psi(f_n + \epsilon z_n^*) - \psi(f_n) \right\} = \frac{\psi(f_n + \epsilon_n \mathbb{Z}_n^*) - \psi(f_n)}{\epsilon_n} \leadsto \psi^{\prime}_{f_\star}(\mathbb{G}_0).
\end{equation*}

Similar to the previous proof, note that $\sqrt{n}\{\psi(f_n)-\psi(f_\star)\} \leadsto \psi_{f_\star}^\prime(\mathbb{G}_0)$, we have
\begin{align*}
    &\sup_{g\in \operatorname{BL}_1 (\mathbb{R})} \Big|\mathbb{E}\left[ g\left(\epsilon^{-1}\left\{\psi(f_n + \epsilon z_n^*) - \psi(f_n) \right\}\right) | f_n\right] -  \mathbb{E}\left[g\left(\sqrt{n}\left\{\psi(f_n) - \psi(f_\star) \right\} \right) \right] \Big| \\
    \leq &\sup_{g\in \operatorname{BL}_1 (\mathbb{R})} \Big|\mathbb{E}\left[ g\left(\epsilon^{-1}\left\{\psi(f_n + \epsilon z_n^*) - \psi(f_n) \right\}\right) | f_n\right] -  \mathbb{E}\left[g\left(\psi^{\prime}_{f_\star}(\mathbb{G}_0) \right) \right] \Big| \\
    & + \sup_{g\in \operatorname{BL}_1 (\mathbb{R})} \Big| \mathbb{E}\left[g\left(\psi^{\prime}_{f_\star}(\mathbb{G}_0) \right) \right] - \mathbb{E}\left[g\left(\sqrt{n}\left\{\psi(f_n) - \psi(f_\star) \right\} \right) \right] \Big| \\
    &= o_p(1) + o_p(1) = o_p(1)
\end{align*}

by triangle inequality. Hence we complete the proof of Theorem \ref{thm:bootConsistency2}. \hfill $\square$

\vspace{1em}

\emph{Proof of Theorem \ref{thm:CITwoSide}.} By standard results in \cite{politis1999subsampling}, under bootstrap consistency, we have
\begin{equation*}
    \liminf_{n \to \infty}  \mathbb{P} \left(\psi(f_\star) \in \left[\psi(f_n) - \frac{c_{1-\alpha/2}^*}{\sqrt{n}}, \psi(f_n) - \frac{c_{\alpha/2}^*}{\sqrt{n}} \right] \right) = 1 - \alpha
\end{equation*}
if the limiting distribution is continuous at the boundary of quantiles;
\begin{equation*}
    \liminf_{n \to \infty}  \mathbb{P} \left(\psi(f_\star) \in \left[\psi(f_n) - \frac{c_{1-\alpha/2}^*}{\sqrt{n}}, \psi(f_n) - \frac{c_{\alpha/2}^*}{\sqrt{n}} \right] \right) > 1 - \alpha
\end{equation*}
if the limiting distribution is discontinuous at the boundary of quantiles. \hfill $\square$

\vspace{1em}

\emph{Proof of Theorem \ref{thm:CIOneSide}.} For any $f_{\star} \in \Delta_K$ such that $\psi(f_\star) \leq \delta$,
\begin{align*}
    & \mathbb{P} \left(\sqrt{n} \psi(f_n) > \sqrt{n}\delta + c_{1-\alpha}\right) \\
    = &1-\mathbb{P} \left(\sqrt{n} \psi(f_n) \leq \sqrt{n}\delta + c_{1-\alpha}\right) \\
    = & 1- \mathbb{P}\left(\sqrt{n}\{\psi(f_n) - \psi(f_\star) \} \leq c_{1-\alpha} + \sqrt{n}(\delta - \psi(f_\star)) \right) \\
    \leq & 1 - \mathbb{P}(\sqrt{n}\{\psi(f_n) - \psi(f_\star)\} \leq c_{1-\alpha}) \\
    \leq & 1 - (1-\alpha)\\
    = & \alpha,
\end{align*}
where $c_{1-\alpha}$ is the $(1-\alpha)$-th quantile of $\sqrt{n}\{\psi(f_n) - \psi(f_\star) \}$.
With Bootstrap consistency,
\begin{align*}
     &\limsup_{n \to \infty} \sup_{f_{\star} \in \Delta_K: \psi(f_\star) \leq \delta} \mathbb{P}_{f_\star} \left(\sqrt{n} \psi(f_n) > \sqrt{n}\delta + c_{1-\alpha}^* \right) \\
     \leq & \limsup_{n \to \infty} \sup_{f_{\star} \in \Delta_K: \psi(f_\star) \leq \delta} \mathbb{P}_{f_\star} \left(\sqrt{n} \psi(f_n) > \sqrt{n}\delta + c_{1-\alpha} \right) = \alpha.
\end{align*}

For any $f_{\star} \in \Delta_K$ such that $\psi(f_\star) > \delta$,
\begin{equation*}
    \mathbb{P} \left(\sqrt{n} \psi(f_n) > \sqrt{n}\delta + c_{1-\alpha}^*\right) \to 1.
\end{equation*} \hfill $\square$

\section{Bootstrap methods}
\label{sec:bootstrapMethods}

\begin{algorithm}
\caption{$m$-out-of-$n$ bootstrap}
\label{alg:m-out-of-n}
\begin{algorithmic}[1]
    \State \textbf{require}: $m$ (rule of thumb: $2\sqrt{n}$), $B\in \mathbb{N}$ 
    \State set $\mathcal{S} = \varnothing$
    \State \textbf{for} $i = 1, 2, \cdots, B$ \textbf{do}:
    \State ~~~~draw $Y^* \sim \operatorname{Multinomial}(m; f_n)$
    \State ~~~~append $\sqrt{m}\{\psi(Y^*/m) - \psi(f_n))\}$ to $\mathcal{S}$
    \State \textbf{end for}
    \State \textbf{output}: $\mathcal{S}$
\end{algorithmic}
\end{algorithm}

\begin{algorithm}
\caption{numerical derivative method}
\label{alg:numerical-bootstrap}
\begin{algorithmic}[1]
    \State \textbf{require}: $\epsilon$ (rule of thumb: $n^{-1/4}$), $B\in \mathbb{N}$ 
    \State set $\mathcal{S} = \varnothing$, $i = 1$
    \State \textbf{while} $i \leq B$ \textbf{do}:
    \State ~~~~draw $Z^* \sim \mathcal{N}(\boldsymbol{0}_K, \Sigma(f_n))$
    \State ~~~~\textbf{if} $f_n + \epsilon Z^* \in \mathbb{R}_+^K$:
    \State ~~~~~~~~append $\epsilon^{-1}\{\psi(f_n+ \epsilon Z^*) - \psi(f_n))\}$ to $\mathcal{S}$
    \State ~~~~~~~~$i \leftarrow i+1$
    \State ~~~~\textbf{else}:
    \State ~~~~~~~~\textbf{continue}
    \State \textbf{output}: $\mathcal{S}$
\end{algorithmic}
\end{algorithm}

\end{document}